\documentclass[journal]{IEEEtran}
\usepackage{cite}
\usepackage{amsmath,amssymb,amsfonts}
\usepackage{algorithmic}
\usepackage{algorithm}
\usepackage{graphicx}
\usepackage{textcomp}
\usepackage{xcolor}
\usepackage{booktabs}
\usepackage{multirow}
\usepackage{subcaption}
\usepackage{url}
\usepackage{placeins}

\begin{document}

\title{Reassessing Global Gradient-Norm Imbalance in BLIP Fine-Tuning Across Physical Domains}

\author{Kiran~Naseer,
        Samreen~Azhar,
        and~Dwarikanath~Mahapatra%
\thanks{K. Naseer (corresponding author, email:
25016119-003@uog.edu.pk) and S. Azhar are with the Department of
Computer Science, University of Gujrat, Pakistan.}%
\thanks{D. Mahapatra is with the Department of Computer Science,
Khalifa University, Abu Dhabi, UAE.}%
\thanks{Manuscript submitted \today.}%
\thanks{Code, training logs, and analysis scripts reproducing all
tables and figures are available at
\protect\url{https://github.com/KiranNaseer-AI/multimodal-domain-shift-vlm}.}}

\maketitle

\begin{center}
\footnotesize
This work has been submitted to the IEEE for possible publication.
Copyright may be transferred without notice, after which this
version may no longer be accessible.
\end{center}

\begin{abstract}
Imbalanced gradient magnitudes between the visual and language
pathways of a vision-language model are often treated as a defect
to be corrected. We test that premise for one family of correction, deliberately excluding adaptive, signal-driven schemes (e.g. BalGrad, OGM, PMR, CGGM), which are a mechanistically distinct class outside this study's
scope. Measuring the language-to-visual gradient-norm ratio,
reported in parameter-normalised form, across nine fine-tuning
conditions, three seeds, and three captioning datasets spanning
distinct physical domain shifts -- underwater, aerial, radiological -- we find imbalance magnitude varies markedly across domains with no predictable ordering. A plain learning-rate reduction cuts imbalance substantially and lands within a few BLEU points of the best method on every dataset.
Staged freezing reduces the ratio on every domain yet never ranks
first; a schedule-only control isolates freezing as the cause on
one dataset but not the other two. Forcing the two gradient groups
to equal magnitude drives per-parameter imbalance close to zero on
every domain, yet is both the best result in the study and the
worst placement among full fine-tuning methods, on different
datasets, with identical settings. Reductions in gradient-norm
ratio do not consistently predict captioning performance across
domains, and how a given level of balance is reached matters as
much as the level itself. As a secondary finding, a commonly reused LoRA configuration applied to BLIP silently adapts zero visual parameters; correcting it improves BLEU-4 on all three datasets.
\end{abstract}


\section{Introduction}

Vision-language models are pretrained on web-scale image--text
corpora and then fine-tuned for a target domain. When that domain
differs from pretraining only in style, fine-tuning is reliable. The harder case is when it differs \emph{physically}: underwater
photography loses wavelength-dependent light, satellite imagery is
shot from nadir at a scale that removes ground-level cues, and
radiology encodes anatomy through intensity rather than natural scene statistics. In each case the visual encoder must learn a very different mapping from pixels to meaning, while the decoder's job -- producing fluent text -- barely changes.

This asymmetry shows up in training as a decoupling between loss
and task metrics on a substantial minority of trained conditions:
downstream captioning quality stagnates or degrades after an
early peak while validation loss remains flat or continues to
fall. Across all nine methods, three datasets and three seeds
(Supplementary Section I-B), we observe this pattern in
34 of 81 runs (42\%), most often on UICD (15/27) and least often
on RSICD (8/27; Supplementary Fig. S6 shows a representative
case where BLEU-4 tracks the loss curve more closely). It is not
universal, and we do not claim it as such; where it occurs, it
coincides with a measurable imbalance between the gradient
magnitudes reaching the visual encoder and the language decoder
during joint training.

A substantial body of work treats this imbalance as a defect and
proposes machinery to correct it, ranging from gradient
modulation in multimodal classification~\cite{peng2022ogm,li2023agm,fan2023pmr,guo2024cggm}
to reweighting schemes aimed specifically at vision-language
models~\cite{gim2025seesaw,paretolora2026,li2025mbq}. The
implicit premise is that correcting the imbalance is the lever
that improves outcomes. This paper tests that premise rather than
adding to the machinery.

We evaluate nine trained conditions against a zero-shot baseline on underwater (UICD), aerial (RSICD) and radiological (ROCOv2)
captioning, three seeds per cell. Alongside standard baselines we
include three conditions that separate mechanisms usually bundled
together: a staged-freezing protocol (Staged FT), a control applying its learning-rate schedule without freezing, and a control forcing the two gradient groups to exact equality every step. Comparing them isolates whether effects come from the schedule, the freezing, or the balancing.

\begin{itemize}
     \item \textbf{Reductions in the language-to-visual
    gradient-norm ratio do not consistently predict captioning
    performance.} Ranking the five full fine-tuning conditions by
    proximity to per-parameter balance correlates only weakly with
    BLEU-4, at Spearman $\rho = -0.30$ on each dataset ($p=0.624$,
    $n{=}5$); the method reducing imbalance most wins on one dataset and loses on the other two.

    \item \textbf{The mechanism by which balance is reached
    matters more than the level reached.} Removing staged freezing while holding the learning-rate schedule fixed collapses the imbalance trajectory back onto the scalar baseline's on UICD (not on RSICD or ROCOv2), and forcing equal group norms at every step gives the best result in the study on one dataset and the worst full fine-tuning result on another, under identical settings.

    \item \textbf{A commonly reused LoRA configuration can
    silently disable visual adaptation on BLIP.} The
    configuration \texttt{target\_modules=[query, value]} matches
    no module in BLIP's vision encoder, adapting zero visual
    parameters. Adding the corresponding vision-side names improves BLEU-4 on all three datasets.
\end{itemize}
All nine trained conditions are non-adaptive. We do not implement or compare against adaptive, signal-driven methods such as
BalGrad~\cite{gim2025seesaw} or OGM~\cite{peng2022ogm,wei2024onthefly};
Section~\ref{sec:scope-adaptive} positions that boundary.
\section{Related Work}
\label{sec:related}

\subsection{Vision-Language Model Fine-Tuning}
Modern captioning models such as BLIP~\cite{li2022blip} and
BLIP-2~\cite{li2023blip2} pretrain a visual encoder and language
decoder jointly, then rely on fine-tuning for downstream adaptation. We use BLIP as our testbed because both its encoder and decoder are fully trainable, which lets us measure and manipulate gradient dynamics between the pathways directly.

\subsection{Staged and Curriculum Training in Multimodal Models}
Staged training -- fine-tuning a subset of a multimodal model's
parameters before unfreezing the remainder -- is an established
pattern: LLaVA~\cite{liu2023llava} and
LLaVA-Med~\cite{li2023llavamed} both freeze a subset of parameters
before unfreezing the remainder. In each case the decision of which parameters to freeze, and when to unfreeze them, is a fixed
architectural choice made once, in advance.

Staged FT as we describe it in Section~\ref{sec:method} follows
the same pattern: which parameters are frozen and for how long is
fixed in advance (two epochs, then three), not adjusted in
response to any measured quantity during training. $R_t$ is
logged throughout but does not influence the training loop. Our
central empirical contribution, however, is not that this distinction produces uniformly better fine-tuning outcomes. It is that it does not: measuring $R_t$ and correcting it structurally proves not to be the causally sufficient intervention the framing implicitly suggests.

\subsection{Parameter-Efficient Fine-Tuning}
LoRA~\cite{hu2022lora} injects low-rank trainable matrices into a
subset of linear layers -- typically the query and value projections -- while freezing the rest. This configuration is standard for language models and is commonly applied unmodified to VLMs. PLoP~\cite{hayou2025plop} has begun to catalogue how sensitive LoRA placement is in multimodal architectures. Our finding in Section~\ref{sec:lora-finding} is a specific instance of that placement-sensitivity problem:
\texttt{target\_modules=[query, value]}, applied verbatim to BLIP,
adapts zero parameters of the visual encoder.

\subsection{Multimodal Imbalance and Gradient Modulation}
\label{sec:related-gradient}
The observation that different modalities in a jointly trained
multimodal model receive imbalanced gradient signal is not new to
this work, and we do not claim it as such. It is an established
research area with a dedicated survey and benchmark
~\cite{balancebenchmark2025}. The canonical diagnosis is that a
modality carrying more readily discriminative information comes
to dominate the joint objective, lowering the loss faster and
thereby suppressing the gradient available to the other
modality~\cite{peng2022ogm,wei2024onthefly}.

A substantial family of methods addresses this by modulating
gradients during optimization, using discriminative
discrepancy~\cite{peng2022ogm,wei2024onthefly}, Shapley-value
attribution~\cite{li2023agm}, class
prototypes~\cite{fan2023pmr}, or classifier-derived
signals~\cite{guo2024cggm} to attenuate the dominant modality.
Within vision-language models specifically, Li et
al.~\cite{li2025mbq} report that the average gradient magnitude of
language tokens exceeds that of vision tokens by roughly an order
of magnitude, motivating modality-aware quantization. See-Saw
Modality Balance~\cite{gim2025seesaw} measures this class of imbalance in vision-language settings and proposes \emph{BalGrad}, correcting it through continuous per-step gradient reweighting and projection. Pareto LoRA~\cite{paretolora2026} reports gradient-magnitude ratios differing by orders of magnitude across tasks and layers in unified multimodal models and applies Pareto-optimal gradient integration.

Two observations position our contribution. First, the survey of this area identifies extension to large pretrained foundation models as an open direction~\cite{balancebenchmark2025}, and our regime sits in that gap. Second, this literature is overwhelmingly
\emph{constructive}: it proposes balancing methods and reports their gains, without systematically asking whether balancing is uniformly beneficial, or comparing mechanistically distinct routes to balance against a properly matched unbalanced control. This is not entirely unremarked: Wei et al. \cite{wei2025improving} argue against balance as an optimality target on bias-variance grounds and report balance-correction methods helping on some datasets while hurting others. Our contribution is narrower: BLIP-based captioning across three physical domain shifts, matched mechanism controls (Joint-Schedule vs. Staged vs. RMS-Balanced), and the BLIP LoRA placement audit, none of which their classification-focused study addresses. Prior work on component-wise gradient-norm clipping~\cite{gnc2022} raised one relevant caveat -- that staged, top-down freezing schemes ``do not always work out due to the different convergence speeds of different layers/modules'' -- but did not test it at the modality level.

\subsection{Scope Relative to Adaptive Methods}
\label{sec:scope-adaptive}
We restrict our comparison to interventions that do not adapt
during training: a fixed two-stage freezing schedule, a fixed
learning-rate schedule, and a fixed per-step rescaling target.
Adaptive, signal-driven methods -- which decide how much to
intervene using a measured discriminativeness, Shapley, or
classifier signal -- are a mechanistically distinct class and are
not evaluated here; we return to why in
Section~\ref{sec:limitations}.

Modality-level imbalance is also distinct from the task-level gradient conflict addressed by GradNorm~\cite{chen2018gradnorm},
PCGrad~\cite{yu2020pcgrad} and multi-objective
formulations~\cite{sener2018mgda}: both our pathways serve a single captioning objective, so imbalance arises from architecture and data, not competing losses. Our contribution is diagnostic rather than methodological.

\subsection{Fine-Tuning Stability Under Distribution Shift}
\label{sec:related-stability}
That aggressive fine-tuning can degrade a pretrained model's useful structure is well established outside the multimodal setting. Kumar et al.~\cite{kumar2022finetuning} show that full fine-tuning can distort pretrained features and underperform out-of-distribution relative to lighter-touch adaptation, with the gap widening as the shift grows. Discriminative and gradually-unfrozen learning rates were introduced for this reason in language-model
transfer~\cite{howard2018ulmfit}, and robust fine-tuning for
zero-shot VLMs trades adaptation strength against retention of
pretrained behaviour~\cite{wortsman2022robustft}. This is why we
treat a reduced-learning-rate baseline (Low-LR FT) as a first-class comparison rather than a throwaway control: it is the cheapest intervention a practitioner has, and prior work gives good reason to expect it to be competitive under shift. Part of what we ask is whether the modality-aware machinery of
Section~\ref{sec:related-gradient} earns its complexity against it.

\section{Method}
\label{sec:method}

\subsection{Problem Formulation}
Let $f_\theta = g_\phi \circ h_\psi$ denote a vision-language
captioning model with visual encoder $h_\psi$ and language
decoder $g_\phi$, where $\theta = (\psi, \phi)$. Given a target
domain $\mathcal{D}$ whose visual distribution differs
substantially from the pretraining distribution, standard
fine-tuning minimizes the captioning loss
\begin{equation}
\mathcal{L}(\theta) = \mathbb{E}_{(x,y)\sim\mathcal{D}}
\left[-\log p_\theta(y \mid x)\right]
\end{equation}
jointly over $\psi$ and $\phi$. We observe empirically
(Section~\ref{sec:uicd-results}) that under physical domain shift
$\mathcal{L}(\theta)$ falls smoothly while held-out task metrics
plateau after an early epoch or degrade.

Part of this gap is definitional. The loss in Eq.~(1) is teacher
forced, so the decoder always sees the ground-truth prefix, while
BLEU-4, CIDEr and METEOR are computed on free-running beam search. A
model can get better at next-token prediction given a correct prefix
while getting no better, or worse, at writing a caption unaided. Every
comparison between the two curves carries that caveat.

The caveat does not explain the pattern away. Supplementary Table S1
reports validation loss -- itself teacher-forced, so comparable across
epochs -- alongside BLEU-4 for Naive FT and Low-LR FT on UICD, seed
42. Low-LR FT's validation loss falls monotonically across all five
epochs while BLEU-4 peaks at epoch 3 and then drops; ordinary
overfitting predicts the loss should rise once BLEU-4 is already
declining, and it does not, at any epoch. Naive FT is a partial
counter-example in the same table: its loss falls for three epochs
then rises mildly over epochs 4 and 5, exactly the overfitting
signature. We do not claim the pattern holds for every method, and we
report Naive FT's late-training loss increase rather than explain it
away.

We use \emph{loss-metric decoupling} descriptively, for the
observation that the two curves diverge in direction rather than
merely in scale, and we do not claim the divergence is caused by
gradient imbalance. That the divergence coincides with a
measurable imbalance is what motivates measuring the imbalance;
it is not evidence that one produces the other.

\subsection{Measuring Gradient Imbalance}
\label{sec:rt}
Concurrent work~\cite{li2025mbq,gim2025seesaw,paretolora2026}
proposes that loss-metric decoupling arises from imbalanced gradient
magnitudes reaching the two pathways. We adopt this diagnostic but
define it carefully, because the obvious formulation misleads.

\subsubsection{Raw ratio}
At step $t$, with $\mathbf{g}_\psi^{(t)}$ and
$\mathbf{g}_\phi^{(t)}$ the gradients w.r.t.\ visual-encoder and
language-decoder parameters, the quantity we log is
\begin{equation}
r_t = \frac{\lVert \mathbf{g}_\phi^{(t)} \rVert_2}
           {\lVert \mathbf{g}_\psi^{(t)} \rVert_2 + \epsilon},
\qquad \epsilon = 10^{-8}.
\label{eq:rt}
\end{equation}

\subsubsection{Why $r_t = 1$ is not the balance point}
Reading $r_t = 1$ as balanced gradient flow is incorrect. The two
groups hold different numbers of parameters, so equal \emph{total}
norms mean unequal \emph{per-parameter} magnitudes: the raw ratio
conflates how hard a pathway is being updated with how large it is. We
therefore also report a parameter-normalised ratio and its
logarithm,
\begin{equation}
R_t^{\mathrm{norm}} =
\frac{\lVert \mathbf{g}_\psi^{(t)}\rVert_2 / \sqrt{N_\psi}}
     {\lVert \mathbf{g}_\phi^{(t)}\rVert_2 / \sqrt{N_\phi}}
= \frac{1}{r_t}\sqrt{\frac{N_\phi}{N_\psi}},
\qquad
B_t = \log R_t^{\mathrm{norm}},
\label{eq:rtnorm}
\end{equation}
where $N_\psi$ and $N_\phi$ are the two groups' parameter counts.
$R_t^{\mathrm{norm}}$ deliberately inverts $r_t$'s numerator and
denominator so its sign reports which pathway dominates per parameter.
For BLIP $\sqrt{N_\phi/N_\psi} = 1.369$, so per-parameter balance
($B_t = 0$) occurs at $r_t \approx 1.37$, not $r_t = 1$; $B_t < 0$
means a larger per-parameter gradient on the language side, $B_t > 0$
on the visual side. We treat $B_t$ as the primary diagnostic: it is
symmetric about balance and well behaved under averaging, unlike a
ratio, where $\overline{1/x} \neq 1/\bar{x}$. Where we quote
$R_t^{\mathrm{norm}}$ we obtain it as $\exp(\bar{B_t})$ so the two
stay consistent.

\subsubsection{Measurement protocol}
\label{sec:measurement-protocol}
Because $r_t$ depends on where in the optimisation step the gradients
are read, we state the protocol in full.

\emph{Group membership.} Parameters are assigned by name:
\texttt{vision\_model} to the visual group, \texttt{text\_decoder}
to the language group; anything matching neither does not enter $r_t$.
Two consequences follow -- the decoder's cross-attention blocks count
as language parameters despite being the fusion point, and so does the
language-modelling head. Section~\ref{sec:sensitivity-results} checks
this choice at initialisation; we did not verify robustness of
between-method orderings across a full training trajectory, which
would require retraining.

\emph{Scaled versus unscaled gradients.} For every condition except
RMS-Balanced FT, $r_t$ is accumulated by backward hooks and so
observes loss-scaled gradients under mixed precision; RMS-Balanced FT
reads \texttt{.grad} after \texttt{unscale\_()}. The loss scale
multiplies both groups equally, so the ratio is invariant to it and
the two are comparable -- absolute norms would not have been.

\emph{Other details.} No gradient clipping and no gradient
accumulation are used; steps at which either group has no gradient are
skipped, and $r_t$ is logged every ten optimiser steps. Supplementary
Section~II records the remaining logging minutiae.

\emph{Optimiser state.} Staged FT builds a fresh AdamW instance at
Stage 2, discarding the moment estimates from Stage 1 -- a potential
confound for anything attributed to unfreezing. Joint-Schedule FT
(Section~\ref{sec:ablation-controls}) resets its optimiser at the same
phase boundary without freezing anything, which is what lets the
comparison isolate freezing from optimiser restart.

We additionally define an effective-update ratio $U_t$, computed from
AdamW's own moment estimates rather than the raw backward-pass
gradient, and log $r_t$ under an alternate grouping convention in
which decoder cross-attention is assigned to the visual rather than
the language group. Both are defined in Supplementary Section~II and
reported in Section~\ref{sec:sensitivity-results}; both were logged
simultaneously with $r_t$ at no additional training cost.

\subsection{Staged FT: A Structured Intervention to Test}
\label{sec:damf}
Staged FT addresses $R_t$ through staged parameter freezing rather
than per-step reweighting. In \textbf{Stage 1} the decoder $\phi$ is
frozen and only the visual encoder $\psi$ trains, for $E_1$ epochs at
$\eta_1$, letting the visual representation adapt without a
simultaneously moving decoder. In \textbf{Stage 2} all parameters are
unfrozen and trained jointly for $E_2$ epochs at $\eta_2 < \eta_1$,
with $R_t$ logged throughout. We are not advocating Staged FT as a
preferred method; we use it as a well-defined structured
$R_t$-correcting intervention to compare against scalar learning-rate
reduction. The comparison, not the intervention, is the finding.

\subsection{Isolating the Mechanism: Two Additional Controls}
\label{sec:ablation-controls}
Staged FT confounds two design choices that could each, independently, explain its behavior: (i) a two-phase learning-rate schedule ($\eta_1$ for $E_1$ epochs, then $\eta_2 < \eta_1$ for $E_2$ epochs), and (ii) staged parameter freezing (Stage 1 trains only $\psi$;
Stage 2 unfreezes $\phi$). To determine which factor drives Staged
FT's $R_t$ trajectory and downstream performance, we introduce two
additional controls that each isolate one factor.

\textbf{Joint-Schedule FT} runs Staged FT's two-phase schedule but
never freezes anything. If Staged FT's elevated late-training $R_t$
came from the schedule shape alone, this control should reproduce it;
if it comes from the freeze/unfreeze structure, the control should
behave like Low-LR FT.

\textbf{RMS-Balanced FT} tests the opposite hypothesis: that direct,
continuous balancing is what determines outcomes. All parameters train
throughout at the constant $\eta = 10^{-5}$ used by Low-LR FT, and
after every backward pass we rescale the two gradient groups so their
root-mean-square magnitudes are equal:
\begin{equation}
g_\psi \leftarrow g_\psi \cdot \frac{\tau}{\mathrm{rms}(g_\psi) + \epsilon},
\qquad
g_\phi \leftarrow g_\phi \cdot \frac{\tau}{\mathrm{rms}(g_\phi) + \epsilon},
\label{eq:rms-balance}
\end{equation}
where $\tau = \sqrt{(\mathrm{rms}(g_\psi) + \epsilon)(\mathrm{rms}(g_\phi) + \epsilon)}$
is the geometric mean of the two groups' RMS gradient magnitudes at
that step, and $\mathrm{rms}(\cdot)$ denotes the root-mean-square
over all elements in a parameter group's flattened gradient. This
forces balance at every step by construction. We log both the
pre-rescale ratio, for comparability with $R_t$ in every other method,
and the post-rescale ratio as a sanity check. Together with Staged FT
and Low-LR FT, these controls let us attribute effects to schedule
shape (Joint-Schedule vs.\ Low-LR), freezing structure (Staged
vs.\ Joint-Schedule), or direct balancing (RMS-Balanced
vs.\ everything else).

\section{Experimental Setup}
\label{sec:setup}

\subsection{Datasets}
\label{sec:datasets}
We evaluate on three datasets, each representing a distinct
physical domain shift relative to BLIP's web-scale pretraining
distribution.

\textbf{UICD} (Underwater Image Captioning Dataset)~\cite{uicd}
contains 3{,}176 underwater images with human-written captions and
severe wavelength-dependent colour attenuation. We split it 70/15/15
at a fixed seed (42), giving 2{,}223 training, 476 validation and 477
test images, held fixed across training seeds so only training
stochasticity varies. The split is at the image level over a flat list
of photographs with no natural grouping, so it does not carry the
leakage risk of ROCOv2's per-image split over multi-image clinical
studies.

\textbf{RSICD}~\cite{lu2018rsicd} contains aerial imagery --
8{,}734 training, 1{,}094 validation and 1{,}093 test images on the
official split, five captions each. Viewpoint and scale change
substantially, but colour and texture stay closer to natural imagery
than in UICD.

\textbf{ROCOv2}~\cite{ruckert2024rocov2} pairs
radiological imagery -- X-ray, CT, MRI, ultrasound -- with clinical
captions. These images are intensity-based and modality-specific, with
no direct analogue in natural-image pretraining; the zero-shot BLEU-4
of $0.0024$ we measure (Table~\ref{tab:main-results}) is the lowest
of the three, though we do not treat that as a calibrated measure of shift. We use the full dataset on its official partition: 59{,}958 training, 9{,}904 validation and 9{,}927 test images. Unlike the other two it gives a single reference caption per image, a comparability caveat we address in Section~\ref{sec:rocov2-results}. We did not re-partition it and did not verify whether the official split separates studies or patients rather than individual images; any leakage there would be present in our results and in other work using it.

\subsection{Baselines and Conditions}
We compare ten conditions, applied identically across all three
datasets. \emph{Pretrained} evaluates zero-shot BLIP as a lower
reference point.

\textbf{Full fine-tuning.} \emph{Naive FT} trains all parameters
jointly at $\eta = 10^{-4}$ and \emph{Low-LR FT} does the same at
$\eta = 10^{-5}$, the cheapest intervention a practitioner has and,
per Section~\ref{sec:related-stability}, one prior work suggests
should be competitive under shift. \emph{Staged FT} is the structured
intervention of Section~\ref{sec:damf}; \emph{Joint-Schedule FT} and
\emph{RMS-Balanced FT} are the controls of
Section~\ref{sec:ablation-controls}.

\textbf{Single-pathway variants.} \emph{Isolated Visual} freezes the
decoder and trains only the visual encoder for two epochs, matching
Staged FT's Stage-1 budget; it isolates what visual-only adaptation
achieves under a matched compute allowance, not an adaptation ceiling.
\emph{Frozen Vision FT} does the reverse.

\textbf{Parameter-efficient.} \emph{Text-Only LoRA Audit}
~\cite{hu2022lora} applies low-rank adapters ($r=16$,
$\alpha=32$, dropout $=0.05$) to the \texttt{query} and
\texttt{value} projections, the commonly reused
\texttt{target\_modules=[query,value]} configuration in our BLIP
implementation. We call this condition \emph{text-only} because,
as Section~\ref{sec:lora-finding} shows, this configuration
resolves entirely to BLIP's text decoder and
adapts zero visual parameters. This was not a design choice; we
discovered it by auditing the architecture after the results came
in. \emph{LoRA (multimodal)} is identical in every respect except
that \texttt{target\_modules} additionally matches
\texttt{qkv} and \texttt{projection}, the corresponding attention
module names in BLIP's vision transformer, bringing 48 visual
modules into the adapted set.

\subsection{Evaluation Protocol}
We report BLEU-4~\cite{papineni2002bleu},
CIDEr~\cite{vedantam2015cider} and
METEOR~\cite{banerjee2005meteor}, computed with the COCO
toolkit~\cite{chen2015cococaptions} against all available references
per image. For each run we take the epoch with the highest validation
BLEU-4 and report all three metrics at that checkpoint, uniformly
across every method, dataset and seed, and we report the best-to-final
BLEU-4 degradation as a measure of training stability.

\subsection{Multi-Seed Statistical Validation}
Every method is trained under three seeds ($42$, $0$, $123$). All runs
start from the same pretrained BLIP checkpoint, so seeds do not vary
pretrained weights; they vary data shuffling, dropout masks and the
initialisation of newly introduced parameters (LoRA adapters), with
the partition held fixed. We report mean $\pm$ standard deviation
across seeds and assess Staged FT against each baseline with a paired
$t$-test on per-seed best-epoch BLEU-4.

Every $p$-value here is descriptive rather than confirmatory: with
three seeds per cell and many pairwise comparisons, these values
indicate which differences are large relative to seed variation but do
not establish effects. We apply no correction for multiple
comparisons, and no conclusion rests on a significance threshold;
where one method loses to another with $p < 0.05$, that is the
strongest directional evidence in our data, not a demonstrated result.
Test-set data are never used for checkpoint or hyperparameter
selection: the best-epoch criterion is computed from validation-split
BLEU-4 only, and test-split metrics are read once, after that
selection is fixed.

\subsection{Implementation Details}
\label{sec:impl-details}
All experiments fine-tune \texttt{Salesforce/blip-image-\allowbreak
captioning-base}~\cite{li2022blip} using AdamW~\cite{loshchilov2019adamw} (weight decay $0.01$) with mixed-precision training~\cite{micikevicius2018mixed}. All images are resized to $384\times384$ by the BLIP image processor's default configuration, regardless of source resolution or dataset. Staged FT uses $E_1{=}2$ Stage-1 epochs at $\eta_1 = 5\times10^{-5}$ and $E_2{=}3$ Stage-2 epochs at $\eta_2 = 10^{-5}$; Joint-Schedule FT
(Section~\ref{sec:ablation-controls}) uses the identical
two-epoch/three-epoch schedule and learning rates without freezing. RMS-Balanced FT trains for 5 epochs at the constant $\eta = 10^{-5}$ used by Low-LR FT. All non-LoRA baselines train for 5 epochs except Isolated Visual, which trains for 2 epochs to match Staged FT's Stage-1 budget. Training batch size is $16$ across all three datasets, held constant across all methods and seeds. Software versions, GPU allocation, and determinism settings are
recorded in Supplementary Section~II. Caption generation at evaluation time uses beam search with beam width $3$ and no length penalty or repetition penalty (HuggingFace default $\alpha=1.0$); maximum generation length is $30$ tokens for UICD and RSICD and $40$ for
ROCOv2, matching the respective training-time caption truncation
lengths.

\section{Results}
\label{sec:results}

We first report per-dataset outcomes for all nine trained
conditions and the zero-shot baseline,
then turn to the cross-dataset picture in
Section~\ref{sec:cross-dataset} and the mechanism ablations in
Section~\ref{sec:ablations}. Table~\ref{tab:main-results}
consolidates best-epoch BLEU-4 across every method and dataset;
per-dataset CIDEr and METEOR, per-seed stability, and paired
significance tests are reported in the Supplementary Material to keep the main text focused on the comparisons that carry the argument.

\begin{table*}[!htbp]
\centering
\caption{Best-epoch BLEU-4 (mean $\pm$ std over seeds 42, 0, 123)
for all nine trained conditions plus the zero-shot baseline, on
all three datasets. Best per dataset
in bold. ROCOv2 uses a single reference caption per image, so its
absolute magnitudes are not comparable to UICD/RSICD; comparisons
are meaningful within a column, not across columns
(Section~\ref{sec:cross-dataset}). On RSICD the top two conditions
differ by $0.0001$, well inside seed noise; both are bolded. Full
CIDEr, METEOR, stability and significance results appear in the
Supplementary Material.}
\label{tab:main-results}
\begin{tabular}{lccc}
\toprule
\textbf{Method} & \textbf{UICD} & \textbf{RSICD} & \textbf{ROCOv2} \\
\midrule
Pretrained (zero-shot) & 0.0819 & 0.0383 & 0.0024 \\
\midrule
\multicolumn{4}{l}{\emph{Full fine-tuning}} \\
Naive FT            & $0.2420 \pm 0.0208$ & $0.4463 \pm 0.0100$ & $0.0215 \pm 0.0008$ \\
Low-LR FT           & $0.2838 \pm 0.0061$ & $\mathbf{0.4788 \pm 0.0045}$ & $0.0238 \pm 0.0002$ \\
Joint-Schedule FT   & $0.2775 \pm 0.0105$ & $\mathbf{0.4789 \pm 0.0032}$ & $0.0244 \pm 0.0007$ \\
Staged FT           & $0.2920 \pm 0.0122$ & $0.4695 \pm 0.0119$ & $0.0221 \pm 0.0005$ \\
RMS-Balanced FT     & $\mathbf{0.3041 \pm 0.0002}$ & $0.4681 \pm 0.0041$ & $0.0241 \pm 0.0004$ \\
\midrule
\multicolumn{4}{l}{\emph{Single-pathway and parameter-efficient}} \\
Isolated Visual     & $0.1835 \pm 0.0208$ & $0.2492 \pm 0.0097$ & $0.0194 \pm 0.0015$ \\
Frozen Vision       & $0.2489 \pm 0.0194$ & $0.4644 \pm 0.0119$ & $0.0211 \pm 0.0008$ \\
Text-Only LoRA Audit    & $0.1813 \pm 0.0031$ & $0.1753 \pm 0.0074$ & $0.0200 \pm 0.0014$ \\
LoRA (multimodal)   & $0.1973 \pm 0.0075$ & $0.1955 \pm 0.0051$ & $\mathbf{0.0246 \pm 0.0004}$ \\
\bottomrule
\end{tabular}
\end{table*}

\subsection{UICD: Underwater Domain Shift}
\label{sec:uicd-results}
Every fine-tuning method improves substantially over the zero-shot
baseline, which manages only $0.0819$ BLEU-4 -- underwater imagery
is far enough from BLIP's pretraining distribution that the model
has little useful prior. Among full fine-tuning methods,
RMS-Balanced FT is strongest at $0.3041$, followed by Staged FT
at $0.2920$; the scalar Low-LR FT baseline reaches $0.2838$.

Two features of these numbers recur throughout. The spread among the
four full-fine-tuning methods is small relative to seed noise for
several pairs -- Staged FT's advantage over Low-LR FT is not
significant ($p{=}0.396$, ahead on 2 of 3 seeds; Supplementary Table
S6). And RMS-Balanced FT's seed-to-seed variance is remarkably small
($\pm 0.0002$, two orders of magnitude tighter than Staged FT's
$\pm 0.0122$), which we return to in
Section~\ref{sec:ablations}.

Supplementary Fig. S7 shows the loss-metric decoupling that
motivated this study: training loss falls smoothly under Naive FT and
Low-LR FT while BLEU-4 stalls or regresses. For Staged FT, BLEU-4
rises sharply at the Stage 1$\to$2 boundary in that seed; we do not
treat this as evidence that unfreezing caused the rise, only that the
two coincide.

\subsubsection{Gradient imbalance}
\begin{figure}[t]
\centering
\includegraphics[width=0.92\columnwidth]{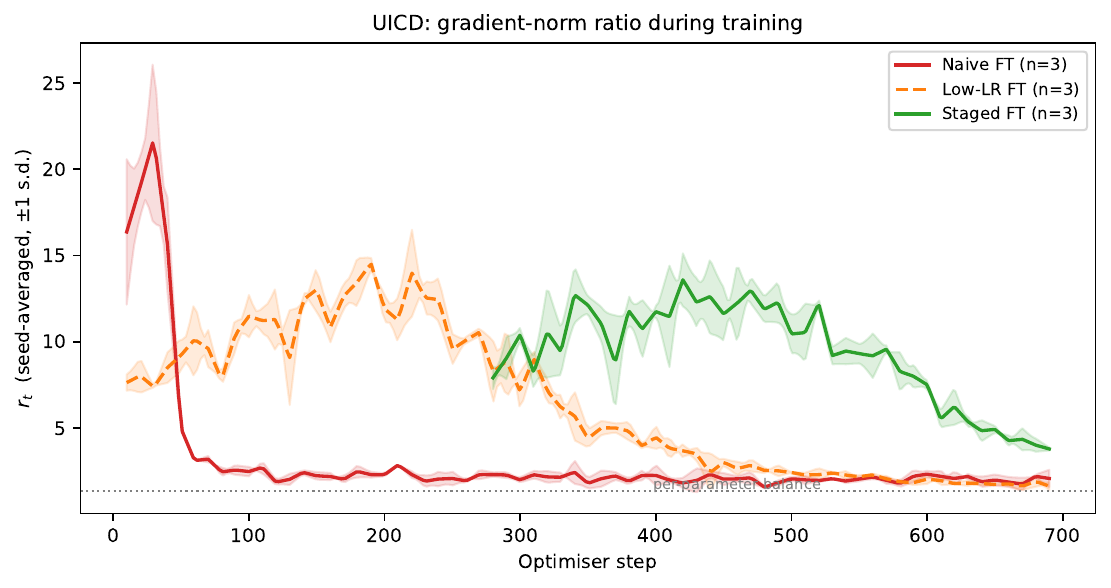}
\caption{UICD. Method-level $R_t$ comparison, seed-averaged.
Na\"ive FT holds a persistently elevated imbalance; Low-LR FT
descends smoothly to a \emph{lower} equilibrium than Staged FT
converges to.}
\label{fig:uicd-rt-comparison}
\end{figure}

Figure~\ref{fig:uicd-rt-comparison} compares $R_t$ trajectories
for three representative methods. Naive FT sits at
$R_t \approx 10.9$ for the whole run with nothing to pull it
down. Low-LR FT starts at a comparable $R_t \approx 9.0$ and
descends smoothly to $\approx 1.8$. Staged FT enters Stage~2
slightly higher ($\approx 10.0$) and settles at $\approx 5.1$ --
noticeably above where the scalar baseline lands. On this
dataset, in other words, the method with the \emph{worse} final
imbalance is the one that scores higher. Per-seed Staged FT
trajectories appear in Supplementary Fig. S4.

\subsection{RSICD: Aerial Domain Shift}
\label{sec:rsicd-results}
RSICD reverses part of the UICD ordering. Joint-Schedule FT and
Low-LR FT are effectively tied at the top ($0.4789$ and $0.4788$
respectively -- a difference well inside seed noise), Staged FT
falls to $0.4695$, and RMS-Balanced FT is last among the
full-fine-tuning methods at $0.4681$. Staged FT loses to Low-LR
FT on all three seeds, though the margin does not reach
significance at $n{=}3$ ($p{=}0.205$; Supplementary Table S10).
The gap between full fine-tuning and the parameter-efficient
conditions is much wider here than on UICD. Both LoRA variants
land near $0.19$--$0.20$ against roughly $0.47$--$0.48$ for full
fine-tuning, and Isolated Visual reaches only $0.2492$. Aerial
captioning appears to need more of the model adapted than a low-rank adapter provides.

\subsubsection{The Stage-2 transient}
The RSICD $R_t$ trajectories expose a mechanism easy to miss when only endpoints are reported. Staged FT does not enter Stage~2 from a corrected state and descend; it enters from a discontinuity: unfreezing the decoder while the visual encoder is
still adapting re-admits the previously suppressed language-side
gradient, and $R_t$ spikes transiently to $15$--$20$ before
decaying to $\approx 1.8$. Averaged over the first ten logged
Stage-2 steps -- already on the rising edge of that spike --
Staged FT's early $R_t$ is $13.57 \pm 0.26$, higher than Low-LR
FT's $11.17 \pm 0.19$ over its own first ten steps. Low-LR FT
reaches its converged $R_t \approx 1.37$ without any such
excursion. Per-seed trajectories are in Supplementary Fig. S5.

\subsection{ROCOv2: Radiological Domain Shift}
\label{sec:rocov2-results}
Absolute scores here are an order of magnitude below the other two
datasets, for two compounding reasons: radiological captioning is
genuinely hard for a naturally-pretrained model, and the single
reference caption per image means BLEU-4 punishes any departure from that one phrasing. We treat magnitudes as comparable only within this dataset.

Within that column, Joint-Schedule FT, LoRA (multimodal) and
RMS-Balanced FT are statistically indistinguishable at $n{=}3$
($0.0244$, $0.0246$, $0.0241$ respectively). Staged FT is
last among full fine-tuning methods at $0.0221$, and its loss to
Low-LR FT is the one comparison in our study that reaches
conventional significance ($p{=}0.042$, losing on all three
seeds; Supplementary Table S12). ROCOv2 is where the pretrained model does worst in absolute terms, subject to the single-reference caveat above, so any account under which structured intervention earns its keep where adaptation is hardest predicts Staged FT doing well here. It does the opposite.

\begin{figure}[t]
\centering
\includegraphics[width=0.92\columnwidth]{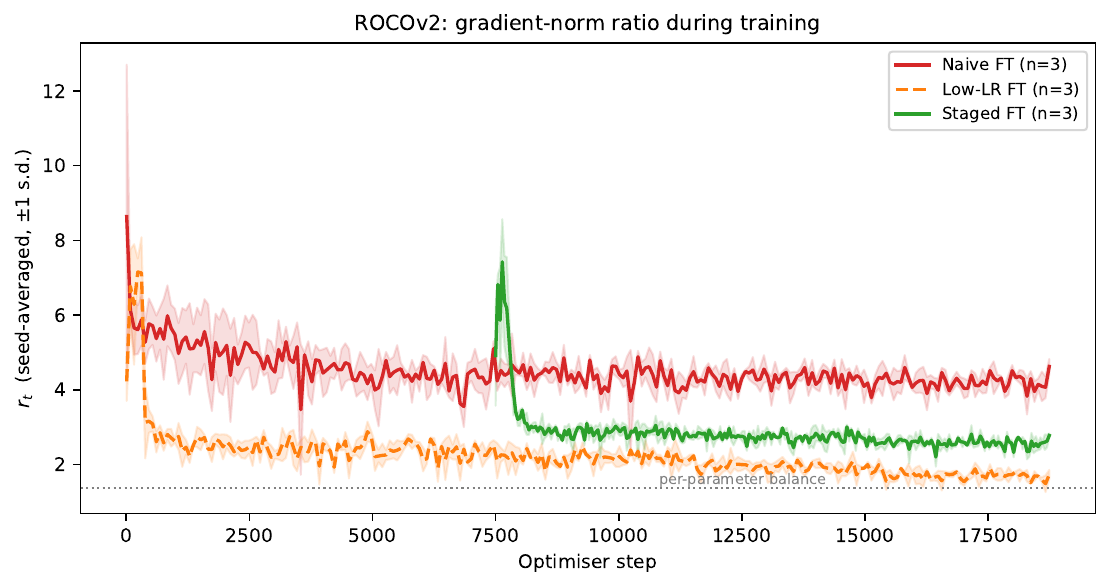}
\caption{ROCOv2, seed-averaged with per-method $\pm 1$ std bands.
Low-LR FT (dashed) descends smoothly to the lowest converged
$R_t$; Staged FT (solid) shows a Stage-2 transient before
settling above it; Na\"ive FT stays persistently elevated. The
same qualitative ordering holds on all three datasets.}
\label{fig:rocov2-rt-comparison}
\end{figure}

The Stage-2 transient reproduces here but more mildly: individual
seeds peak at $R_t \approx 10.0$--$11.3$, averaging $6.10 \pm
0.26$ over the first ten Stage-2 steps against Low-LR FT's
$5.68 \pm 0.41$. That narrower gap is consistent with ROCOv2's
milder baseline imbalance, discussed in
Section~\ref{sec:cross-dataset}.

\subsection{A Semantic Consistency Check}
\label{sec:bertscore}
As a secondary, exploratory check we compute BERTScore F1~\cite{zhang2020bertscore}
on a fixed subsample of predictions retained during training (20 test images per seed, identical image indices across all methods within a dataset); the full table and per-dataset discussion are in Supplementary Section~II. On UICD it separates full-fine-tuning
methods ($0.558$--$0.587$) cleanly from single-pathway and
parameter-efficient methods ($0.480$--$0.545$), agreeing with BLEU-4's broad ordering, and it corroborates our treatment of UICD CIDEr as brittle rather than diagnostic (Supplementary Section I-A): Joint-Schedule FT and RMS-Balanced FT score competitively ($0.571$, $0.569$) despite depressed CIDEr ($0.14$--$0.16$). RSICD and ROCOv2 show no comparable separation and we draw no dataset-level conclusions from BERTScore there. The subsample is smaller than, and independent of, the full test sets used for the primary metrics, and we do not treat it as comparable to them in magnitude.

\section{Cross-Dataset Synthesis}
\label{sec:cross-dataset}

Supplementary Table S14 summarises the diagnostic picture across all three datasets, reporting per (dataset, method) cell the best BLEU-4, the stability degradation, and the $R_t$ trajectory endpoints, with early $R_t$ computed separately for each method to reflect where its trainable window actually begins. Three cross-cutting observations emerge.

\emph{First, $R_t$-early varies by roughly a factor of two across
domains, with no ordering we can predict.} Both methods begin joint training at $R_t \approx 11.2$ on RSICD, $\approx 9.0$ on UICD and $\approx 5.7$ on ROCOv2. We cannot derive that ordering from any independently measurable property of the domains, and we have no calibrated instrument for ranking them by distance from natural-image pretraining.

\emph{Second, Staged FT's Stage-2 boundary $R_t$ is consistently
higher than Low-LR FT's early $R_t$, and so is its late $R_t$, on
every dataset.} Low-LR FT converges to a lower equilibrium everywhere, with no architectural intervention.

\emph{Third, the relationship between $R_t$ dynamics and performance is dataset-dependent rather than universal.} Within a single dataset, a positive association between $R_t$ statistics and BLEU-4 appears on two of three datasets, not RSICD
(Supplementary Section I-C); at $n{=}6$ pooling two methods, it cannot be separated from a difference between the methods themselves, so we report it descriptively. Either way it does not translate into a consistent cross-dataset ranking: Low-LR FT
achieves the larger $R_t$ reduction than Staged FT on all three datasets (UICD $4.99\times$ vs.\ $1.97\times$; RSICD $8.15\times$ vs.\ $7.71\times$; ROCOv2 $3.40\times$ vs.\ $2.42\times$), yet Staged FT still wins on BLEU-4 on UICD. The method making the smaller correction beats the one making the larger. $R_t$ carries some within-dataset signal; it does not tell you which intervention to deploy across domains.

\subsection{Imbalance measured per parameter}
\label{sec:normalised-analysis}

Supplementary Table S14 reports the raw ratio, which is what the
logging captured and what prior work typically quotes. As
Section~\ref{sec:rt} notes, however, raw ratios conflate update
strength with group size. Supplementary Table S15 reports the same runs under the
parameter-normalised signed log-ratio $B_t$, for all five
full-fine-tuning conditions; we summarise what it shows here.

Three observations follow, and they sharpen rather than overturn
the picture from the raw ratios. \emph{First, every condition begins
language-dominated and moves toward balance.} Early $B_t$ lies
between $-1.39$ and $-2.24$ everywhere, meaning the language pathway
receives roughly four to nine times the per-parameter gradient of the
visual pathway at the start of joint training; late $B_t$ moves
toward zero in every case. \emph{Second, forcing equal group norms
produces near-exact per-parameter balance.} RMS-Balanced FT targets
$R_t^{\mathrm{norm}} = 1$ directly rather than the raw ratio
$r_t = 1$, and its late-training $B_t$ sits close to zero on all three
datasets ($-0.04$, $+0.14$, $-0.21$; corresponding raw ratios $1.44$,
$1.24$, $1.70$, all near the per-parameter balance point of $1.37$).
For this method $r_t$ is logged as
$\mathrm{rms}(g_\phi)/\mathrm{rms}(g_\psi)$ rather than the raw
total-norm ratio used elsewhere
(Section~\ref{sec:measurement-protocol}); values here have been
converted to the raw-ratio convention for comparability. Despite
ending nearest per-parameter balance of any method on UICD, its UICD
and RSICD placements reverse entirely
(Section~\ref{sec:ablations}); near-exact balance does not guarantee
good performance. \emph{Third, proximity to balance still does not reliably predict
performance.} Ranking methods within each dataset by
$\lvert B_t \rvert$ and correlating against best BLEU-4 gives
Spearman $\rho = -0.30$ on all three datasets ($p=0.624$, $n=5$) --
identical across datasets because at $n{=}5$ the rank statistic takes
few discrete values. The direction is consistent, the association
small. Nor is it the whole story: RMS-Balanced FT ends closest to
balance on UICD ($B_t = -0.04$) and wins there, but is only
second-closest on RSICD ($+0.14$, behind Low-LR FT's $+0.02$) and
ROCOv2 ($-0.21$, behind $-0.18$), placing near the top on one and last
among full fine-tuning methods on the other. A weak average tendency
toward better performance near balance coexists with concrete
counter-examples per dataset, which is why we treat proximity to
balance as suggestive rather than predictive.

\subsection{Effective-Update and Grouping-Sensitivity Results}
\label{sec:sensitivity-results}

Two methodological choices warrant a direct check rather than
acknowledgment alone: whether RMS-Balanced FT's per-step rescale
(Eq.~\ref{eq:rms-balance}) changes the effective optimisation step
size independent of balancing, and whether our group-membership
convention (Section~\ref{sec:measurement-protocol}) drives any
reported conclusion. Trained checkpoints were not retained, so we
test both at initialisation on a 16-example RSICD batch using the
unmodified pretrained checkpoint; this establishes whether either is
a first-order confound, not whether it holds throughout training.
\emph{Effective update.} Rescaling to equal RMS magnitude changes the
combined gradient norm from $28.96$ to $27.00$, a $6.8\%$ reduction --
directionally toward a smaller effective step at the same nominal
learning rate. Whether that compounds meaningfully over a full run,
plausibly contributing to RMS-Balanced FT's divergent UICD/RSICD
outcomes, cannot be determined from an initialisation-only check.
\emph{Parameter grouping.} Re-assigning cross-attention changes early
$r_t$ from $1.813$ (our convention) to $1.686$ (cross-attention moved
to the visual group) to $1.781$ (excluded from both) -- a difference
of $0.127$ between the two extremes, roughly $7\%$. All three groupings place $r_t$ above the per-parameter balance point of
$1.37$ (Section~\ref{sec:rt}), so the qualitative finding that
pathways begin joint training language-dominated is unchanged under every grouping we tested.

\section{Mechanism Ablations: Isolating Freezing from Scheduling and Balancing}
\label{sec:ablations}

Section~\ref{sec:cross-dataset} shows that Staged FT does not
reliably beat Low-LR FT despite Low-LR FT achieving the larger Rt reduction on every dataset. That leaves an obvious question unanswered. Staged FT bundles two design choices -- a two-phase
learning-rate schedule, and staged parameter freezing -- and
from the results so far there is no way to tell which one is
responsible for its behaviour. The two controls introduced in
Section~\ref{sec:ablation-controls} separate them.
Table~\ref{tab:ablation-four-way} places all four full
fine-tuning methods side by side.

\begin{table}[!htbp]
\centering
\caption{Mechanism ablation. Four full fine-tuning methods that
differ in exactly which factor they apply: a two-phase
learning-rate schedule, staged freezing, or per-step gradient
rescaling. Converged $R_t$ is the mean over the last ten logged
steps. Reading down each
column shows that no single mechanism wins everywhere, and
reading across the Staged FT row shows it never wins anywhere.
Best-epoch BLEU-4 for these four methods is reported in
Table~\ref{tab:main-results} and is not repeated here.}
\label{tab:ablation-four-way}
\begin{tabular}{lcccc}
\toprule
& \multicolumn{2}{c}{\textbf{Factors applied}} & & \\
\cmidrule(lr){2-3}
\textbf{Method} & \textbf{Two-phase LR} & \textbf{Freezing}
& \textbf{Per-step rescaling} & \\
\midrule
Low-LR FT          & no  & no  & no  & \\
Joint-Schedule FT  & yes & no  & no  & \\
Staged FT          & yes & yes & no  & \\
RMS-Balanced FT    & no  & no  & yes & \\
\midrule
\multicolumn{5}{l}{\emph{Converged $R_t$ (mean over final ten logged steps)}} \\
\midrule
\textbf{Method} & \textbf{UICD} & \textbf{RSICD} & \textbf{ROCOv2} & \\
\midrule
Low-LR FT         & $1.81$ & $1.37$ & $1.67$ & \\
Joint-Schedule FT & $1.77$ & $1.82$ & $2.77$ & \\
Staged FT         & $5.09$ & $1.76$ & $2.52$ & \\
RMS-Balanced FT   & $1.44$ & $1.24$ & $1.70$ & \\
\bottomrule
\end{tabular}
\end{table}

\subsection{The Learning-Rate Schedule Is Not the Cause (UICD)}
Joint-Schedule FT runs Staged FT's exact two-phase schedule without
ever freezing a parameter. If the schedule shape produced Staged FT's
elevated $R_t$, this control should reproduce it.

It does not, straightforwardly. Joint-Schedule FT converges to
$r_t = 1.77$ on UICD, $1.82$ on RSICD and $2.77$ on ROCOv2. On UICD
that sits almost exactly on Low-LR FT's $1.81$ and far below Staged
FT's $5.09$, so holding the schedule fixed and removing only the
freezing collapses $R_t$ back toward the unfrozen baseline -- a
controlled result implicating the freeze/unfreeze structure rather
than the learning-rate curve. On RSICD and ROCOv2 the same control
does not isolate the mechanism: Joint-Schedule FT's converged $R_t$
sits at or above Staged FT's own ($1.82$ vs.\ $1.76$; $2.77$
vs.\ $2.52$), so removing the freezing there does not lower the
imbalance. We therefore attribute Staged FT's elevated equilibrium to
freezing on UICD only, and read the other two datasets as evidence
that the schedule alone can independently produce an elevated
equilibrium.

Downstream the same control tells a second story. Joint-Schedule FT
and Low-LR FT are within seed noise on RSICD ($0.4789$
vs.\ $0.4788$) and close on ROCOv2 ($0.0244$ vs.\ $0.0238$); on UICD
it is the weaker of the two ($0.2775$ vs.\ $0.2838$). The two-phase
schedule buys essentially nothing on its own, which makes Staged FT's
underperformance harder to attribute to anything but the freezing.

\subsection{Forcing balance directly is not uniformly good either}
RMS-Balanced FT takes the opposite approach: no freezing, no
schedule, just a per-step rescaling that equalises the visual and
language gradient groups by construction
(Eq.~\ref{eq:rms-balance}). The post-rescale RMS ratio
$\mathrm{rms}(g_\phi)/\mathrm{rms}(g_\psi)$ sits at exactly
$1.000$ at every logged step on all three datasets, confirming
the rescale operation does what it is designed to do. This is
distinct from the raw total-norm ratio $r_t$ reported elsewhere
in this paper, which -- once converted to the same convention --
converges to $1.44$, $1.24$ and $1.70$
(Section~\ref{sec:cross-dataset}).

The outcome is not what a straightforward reading of the
imbalance literature would predict. On UICD, forcing balance
produces the best result in the study, $0.3041$, ahead of Staged
FT's $0.2920$ and Low-LR FT's $0.2838$ -- and it does so with a
seed-to-seed standard deviation of $0.0002$, roughly sixty times
tighter than Staged FT's $0.0122$ on the same dataset. On ROCOv2
it is again near the top, $0.0241$, statistically
indistinguishable from Joint-Schedule FT's $0.0244$ at $n{=}3$.

On RSICD it is last of the four, at $0.4681$.

That reversal is the most informative single result in the
study, so it is worth stating plainly: the same intervention,
applied identically, is the best available option on one dataset
and the worst on another. Forcing $R_t$ to unity is not a
uniformly good thing to do.

Low-LR FT's natural converged $R_t$ differs by dataset ($1.37$ on
RSICD, $1.67$ on ROCOv2, $1.81$ on UICD), and forced balancing helps
most where that unforced equilibrium sits furthest from unity. We
return to what that might mean in Section~\ref{sec:discussion}.

\subsection{What the ablations establish}
Three things follow from Table~\ref{tab:ablation-four-way} read alongside Table~\ref{tab:main-results}.

First, the freezing attribution above holds only on UICD.

Second, Staged FT never wins: it places second on UICD, third on
RSICD, fourth on ROCOv2. Against only Low-LR FT
(Section~\ref{sec:cross-dataset}) it appeared to have one home
dataset; with the controls included even that disappears, because
RMS-Balanced FT beats it on UICD by a margin well outside seed
variance.

Third, no alternative dominates either. RMS-Balanced FT wins UICD and
near-ties ROCOv2 but is last on RSICD; Joint-Schedule FT leads RSICD
and ROCOv2 but is last on UICD. Whichever mechanism one picks, some
dataset in our study makes it the wrong choice.

\section{Discussion}
\label{sec:discussion}

\subsection{What freezing actually does}
Staged FT rests on an appealing premise: let the visual encoder
realign in isolation, then unfreeze and adapt both pathways from a
corrected starting point. Our trajectories suggest Stage~2 does not
begin from a corrected state at all, but from a discontinuity.
Unfreezing the decoder while the encoder is still moving re-admits the language-side gradient Stage~1 suppressed, and $R_t$ spikes -- to $15$--$20$ on RSICD, $10$--$11$ on ROCOv2 -- before decaying (Supplementary Figs. S5 and S3). Low-LR FT never absorbs such an excursion; it descends smoothly from initialisation. We describe this as an observed sequence, not a demonstrated causal chain: the optimiser reset at the stage boundary
(Section~\ref{sec:measurement-protocol}) coincides with the
unfreezing, and the Joint-Schedule control confirms the attribution on UICD only (Section~\ref{sec:ablations}). Whether the round trip is worth taking is dataset-dependent, and on our three datasets the answer is never clearly yes.

\subsection{Balance is not a target to be maximised}
That RMS-Balanced FT does not help everywhere argues against reading $R_t$ as a quantity to be minimised. A more defensible reading is that the ratio a well-behaved optimiser settles into carries information about the domain -- about how much of the adaptation burden genuinely
falls on each pathway for that kind of image and caption. Overriding
it substitutes a fixed prior for that signal, and where the natural
equilibrium was already near unity, as on RSICD, there was little to
gain and something to lose. We do not want to overstate this: three
datasets cannot establish a rule, and we did not sweep the balance
target, which would be the direct test. But the pattern matches
Section~\ref{sec:cross-dataset}, where any within-dataset association fails to rank methods across datasets.

\subsection{A silent failure mode in a commonly reused configuration}
\label{sec:lora-finding}
LoRA's weak showing prompted us to audit which modules were
actually being adapted, and the answer was instructive.

The \texttt{target\_modules=[query, value]} configuration matches
by leaf module name. In BLIP, the text decoder's attention uses
separate \texttt{query}, \texttt{key} and \texttt{value} linear
layers, so those names match. The vision transformer does not: it
fuses the three projections into a single \texttt{qkv} layer,
with output projection named \texttt{projection}. Neither string
appears in the default pattern.

The consequence is that the default configuration resolves to 48
adapted modules, all of them in the text decoder, and zero in the
visual encoder, with no warning: training proceeds, loss decreases, and the adapter cannot change the pathway a physical domain shift most requires changing.

Adding \texttt{qkv} and \texttt{projection} brings 48 visual
modules into the adapted set, with rank, scaling, learning rate and seeds unchanged. BLEU-4 improves on every dataset: $+0.0160$ on UICD, $+0.0202$ on RSICD, $+0.0046$ on ROCOv2. Two
qualifications matter. This is a property of BLIP's naming convention rather than a flaw in LoRA or PEFT, and a specific instance of the placement-sensitivity problem that~\cite{hayou2025plop} addresses systematically. And the fix recovers performance that was being discarded silently, not parity with full fine-tuning.

\subsection{Limitations}
\label{sec:limitations}

Five limitations bear naming directly.

\textbf{This study is scoped to non-adaptive interventions by
design, not by gap.} RMS-Balanced FT forces exact per-step
equality and nothing more; it is not a reimplementation of
BalGrad~\cite{gim2025seesaw}, OGM~\cite{peng2022ogm,wei2024onthefly},
or Pareto LoRA~\cite{paretolora2026}, all of which modulate
adaptively using richer signals, and we did not attempt to make it
one (Section~\ref{sec:scope-adaptive}). Forcing exact equality every step is a specific RMS-rescaling intervention, not a clean isolation of balance in the abstract — the rescale also changes the combined gradient magnitude and AdamW trajectory (next paragraph). Its dataset-dependent outcome is evidence that the gradient-ratio statistic alone does not reliably rank the tested methods, not a verdict on balance itself. A dedicated comparison against adaptive reweighting is a natural next study rather than an omission from this one.

Second, the RMS-Balanced FT rescale (Eq.~\ref{eq:rms-balance})
sets each group's magnitude to their geometric mean $\tau$, which
is not constructed to preserve the combined pre-rescale gradient
norm. The method's nominal learning rate therefore does not
necessarily correspond to the same effective step size as
Low-LR FT's, and part of the performance difference between them
could reflect this rather than the balancing itself. A norm-preserving variant would isolate this. Section ~\ref{sec:sensitivity-results} reports an initialisation-only first-order check of this and the parameter-grouping question; neither check extends across a full training trajectory.

\textbf{Three seeds is a weak basis for significance testing.}
We report exact $p$-values rather than thresholding, and we treat
directional consistency across seeds as complementary evidence,
but several of our comparisons are underpowered. The one result
that reaches conventional significance (Staged FT losing to
Low-LR FT on ROCOv2, $p{=}0.042$) should be read alongside the
many that do not.

\textbf{The within-dataset $R_t$--BLEU-4 correlations pool two
methods.} The coefficients in Supplementary Section I-C combine
Low-LR FT and Staged FT at $n{=}6$ per dataset and cannot separate a genuine $R_t$-performance relationship from an intercept difference between the two methods; they are distinct from the $\lvert B_t \rvert$ ranking correlation of
Section~\ref{sec:normalised-analysis}, which ranks five methods at
$n{=}5$.

\textbf{One model family.} All results are on BLIP-base.
Whether the mechanism findings transfer to BLIP-2's frozen-encoder plus Q-Former design, or to decoder-heavy architectures such as LLaVA, is untested. The LoRA naming finding is explicitly
BLIP-specific; the $R_t$ findings we would expect to be more
general but have not verified.

\section{Conclusion}

We set out to reassess a premise much recent work takes as given: that
gradient imbalance between the visual and language pathways is a
defect worth correcting, and that correcting it improves fine-tuning
under domain shift. Across three physically distinct domains, nine
trained conditions and three seeds each, the premise turns out to be
too simple in two ways.

The first is that the mechanism matters more than the correction.
Staged freezing, per-step rescaling and a plain learning-rate
reduction all reduce $R_t$, yet produce different outcomes, and the
differences are not small. Staged freezing was the weakest of the
three: it never ranked first on any dataset. On UICD, the one dataset
where we can isolate the mechanism cleanly, removing only the freezing
while keeping the schedule shows that freezing is what leaves it at an
elevated equilibrium; on the other two datasets the same control does
not disambiguate, and we do not extend the attribution there.

The second is that balance is not something to maximise. Forcing
$R_t$ to exactly unity at every step gave us both the single best
result in the study and the single worst placement among full
fine-tuning methods, on different datasets, with identical
settings. Whatever the ratio an unforced optimiser settles into
represents, overriding it is not reliably an improvement.

Two smaller findings may be of practical use. Initial imbalance varies by roughly a factor of two across our three domains with no ordering we could predict in advance, so it has to be measured rather than assumed. And the commonly reused
\texttt{target\_modules=[query,value]} LoRA configuration, applied
unmodified to BLIP, adapts zero parameters of the visual encoder
without any warning; correcting the target module list recovers real performance on all three datasets, though not enough to close the gap to full fine-tuning.

This study was scoped to non-adaptive interventions; applying the same matched-control logic to an adaptive method is a natural next step. For a practitioner choosing among the options tested here, a reduced learning rate came within a few BLEU points of the best method on every dataset, at no implementation cost and no additional hyperparameters.

\FloatBarrier
\footnotesize
\bibliographystyle{IEEEtran}
\bibliography{references}
\end{document}